# Transformer-based unsupervised patient representation learning based on medical claims for risk stratification and analysis

Xianlong Zeng
Electrical Engineering and
Computer Science
Ohio University
Athens Ohio USA
xz926813@ohio.edu

Simon Lin
Research Information Solutions
and Innovation
Nationwide Children's Hospital
Columbus Ohio USA
Simon.Lin@nationwidechildrens.org

Chang Liu
Electrical Engineering and
Computer Science
Ohio University
Athens Ohio USA
liuc@ohio.edu

## ABSTRACT

The claims data, containing medical codes, services information, and incurred expenditure, can be a good resource for estimating an individual's health condition and medical risk level. In this study, we developed Transformer-based Multimodal AutoEncoder (TMAE), an unsupervised learning framework that can learn efficient patient representation by encoding meaningful information from the claims data. TMAE is motivated by the practical needs in healthcare to stratify patients into different risk levels for improving care delivery and management. Compared to previous approaches, TMAE is able to 1) model inpatient, outpatient, and medication claims collectively, 2) handle irregular time intervals between medical events, 3) alleviate the sparsity issue of the rare medical codes, and 4) incorporate medical expenditure information. We trained TMAE using a real-world pediatric claims dataset containing more than 600,000 patients and compared its performance with various approaches in two clustering tasks. Experimental results demonstrate that TMAE has superior performance compared to all baselines. Multiple downstream applications are also conducted to illustrate the effectiveness of our framework. The promising results confirm that the TMAE framework is scalable to large claims data and is able to generate efficient patient embeddings for risk stratification and analysis.

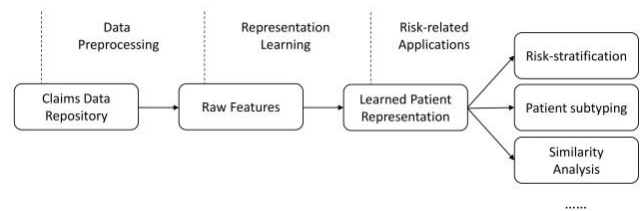

**Figure 1. Our approach toward patient risk stratification and analysis using unsupervised representation learning on claims data.**

## 1 Introduction

The healthcare system in the United States is highly inefficient and costly: rapidly growing healthcare expenditures have become one of the most significant challenges to healthcare providers and care organizations [1]. Unless the medical cost growth is kept in check, the healthcare system might become unsustainable [2]. One possible way to mitigate this arising cost is to pay more attention to preventive care instead of devoting substantial resources to acute medical condition treatment [3]. According to [4], more than $30 billion in medical costs were preventable in 2006. Public health interventions heavily rely on the effective stratification of the entire population into different risk groups and then manage them differently.

Traditional risk-stratification approaches rely on medical experts to handcraft relevant features. For example, the Diagnostic Cost Groups (DCG) model [5] first leverages domain experts to manually group thousands of different diseases into semantic-similar categories and then assign risk coefficients to each category via linear regression. Adjusted Clinical Groups (ACG) [6] utilize the human-developed 34 Aggregated Diagnostic Groups (ADGs) to predict the medical risk of beneficiaries from different medical plans. Although these models help assess patients' risk levels, they have the following drawbacks: First, they are labor-intensive to develop and maintain updates. Second, there may be biases toward specific populations as risk assignments might vary across different populations.

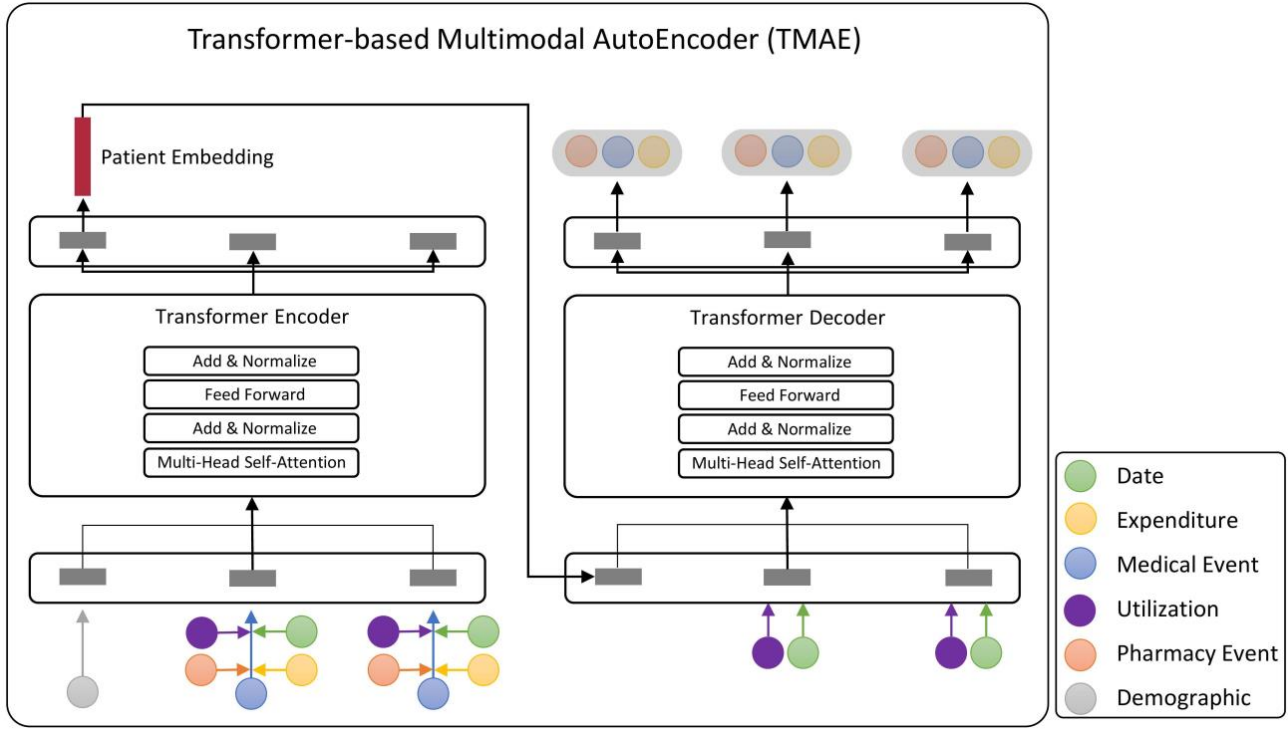

**Figure 2: Framework of our proposed Transformer-based Multimodal AutoEncoder (TMAE). TMAE takes sequential multi-modal medical visits as inputs, including demographics, diagnosis codes, procedure codes, medication codes, claims type, expenditure, and service date. Since a medical visit might consist of multiple codes, the visit embedding dimensions might vary across the timeline. A pooling layer is applied to address this issue and helps to extract the salient information of each visit. The state-of-art transformer blocks are adopted to encode information for patient representation learning. The learned patient embeddings are forced to encode the clinical and resource usage information by the decoder, which reconstructs the medical & pharmacy events and the corresponding expenditures. A multi-objective joint training scheme is proposed to enable efficient training.**

The increasing adoption of Electronic Health Records (EHR) and claims data provides us an opportunity to perform risk stratification in a data-driven manner instead of heavily relying on expert domain knowledge. Researchers first build predictive models to model patients' future medical expenditures and stratify patients according to the predicted cost. Bertsimas et al. [7] utilize tree-based machine learning models to model medical costs using historical medical codes. Zeng et al. [8] develop an interpretable sequential deep learning framework to model next year's medical expenditure. Despite the supervised approach being straightforward, they do not perform very well in the real-world scenario. A major challenge is to find the appropriate outcome label. Many studies utilized the incurred medical cost as the outcome variable, but costs can be both temporary, and neither does it reflect the clinical severity [9]. A large portion of the high-cost visits cannot be predicted based on the historical medical records, leading to a noisy input & output combination for the predictive models. We believe achieving significantly better future expenditure prediction results require additional personal and environmental features, such as financial status, location of residence, means of transportation, lifestyle, and so forth. Since such information is rarely included in the claims data, supervised approaches toward risk stratification are challenging.

Motivated by the practical need to risk-stratify the patient population and understand the characteristics of the subgroups of enrollees, we seek to develop an unsupervised patient representation learning framework to encode the complex claims data for stratification and analysis. Our work is inspired by several observations. First, heterogeneity among patients leads to subtypes within a disease-specific population. These subgroups of patients differ in disease progression and utilization patterns. Therefore, if we can derive disease subtypes, we have an opportunity to guide care organizations for better care management and efficient care interventions. Second, from the computational perspective, risk stratification and patient subtyping can be formulated as clustering tasks. The key is to learn an efficient representation of the current knowledge about a patient in a way that could summarize the medical, utilization, and expenditure information. Figure 1 presents the pipeline of our proposed approach to tackle the risk stratification task. However, to encode the complex information within claims data for efficient patient representations, challenges remain:

- Medical codes (i.e., diagnosis, procedures, and medications), utilizations, and incurred costs need to be modeled collectively.
- Medical visits are unevenly distributed over time. The time spans between two successive visits carry important information and need to be considered.
- Some rare medical codes, such as pulmonary fibrosis, suffer from severe sparsity.
- Visits with higher medical expenditure often indicate severe health problems and therefore should have higher representative weight.

To overcome these challenges for learning effective patient representation, we developed a tailored Transformer-based Multimodal AutoEncoder (TMAE) to model the claims data. Our TMAE follows the classic encoder-decoder framework with a bottleneck layer for patient representation learning. We adopt the powerful transformer architecture [10] as the building block, which has been shown to perform well in encoding multimodal information [11]. The general framework of our model is shown in Figure 2. The model can be divided into two parts, i.e., the stacked transformer encoder part and the stacked transformer decoder part. Medical codes, utilization, expenditure, date within a visit are first converted into a binary vector and then fed through an embedding layer. Next, the pooling is applied and aggregate these embeddings. The resulting vector is fed through the encoder and yields a patient embedding. Finally, the decoder tries to reconstruct the medical code and medical expenditure information based on the date, utilization, and patient embedding.

In summary, the main contributions of our work are listed below:

- We proposed a novel unsupervised patient representation learning framework tailored for medical claims data. The model can encode various claims collectively and generate effective embedding for risk stratification and analysis.
- We carefully designed the neural network architecture to handle the irregular time intervals between claims and mitigate the sparsity issue of rare medical events by incorporating the existing medical domain knowledge.

## 2 Related Work

For the purposes of this study, we differentiate the patient risk stratification methods into two different categories: 1) supervised risk prediction models and 2) unsupervised patient subtyping models. We discuss these related works in the following two subsections in detail.

### 2.1 Supervised Risk Prediction Models

Most of the supervised risk prediction models are conducted in a retrospective manner. The general idea is to use historical year-two (y2)'s data as the label and build predictive models based on year-one (y1)'s data as input. Traditional rule-based models mainly rely on domain experts to group thousands of diagnostic medical codes into hundreds of categories. Adjusted Clinical Groups (ACG) [6] and Diagnostic Cost Groups (DCG) [5] were developed to predict medical risk based on diagnostic data. The authors collected patients' medical claims from health maintenance organizations (HMOs). Regression models were applied to the hand-crafted medical categories to predict the morbidity burden. Chronic Disease Score (CDS) [12] was developed to predict future medical costs and hospital visits. They collected 250,000 managed-care enrollees from age 18 and above and utilized different types of medication within six months to calculate the final score. This risk score is a combined measurement of drug usage, which has a strong indication of chronic conditions suggested by medical experts. The Medicaid Rx model [13] was developed for the Medicaid-insured population and utilized the demographic and pharmacy data to adjust per-person payment toward healthcare plans. The Medicaid Rx model can be seen as a refinement of the CDS.

More recently, deep learning approaches were proposed to model complex medical information and learn a good prediction function. Among various neural networks, Recurrent Neural Networks (RNNs) are particularly popular due to their ability to capture the sequential nature of claims data. Choi et al. proposed DoctorAI [14] to model the electronic health records for future diagnosis risk prediction. GRAM [15] and PRIME [16] were developed to incorporate medical domain knowledge for estimating patients' heart failure risk based on their prior medical visits. Zeng et al. [17] and Morid et al. [18] proposed a sequential deep learning model to predict next year's medical cost for estimating a patient's risk level. Yang et al. [19] evaluate the explainability and fidelity of the current deep learning models with respect to predicting future medical costs. All these models treat risk stratification as a supervised task and use the future expenditure or diagnosis as the label.

### 2.1 Unsupervised Patient Embedding Learning

Patient risk stratification can also be viewed as a subtyping task from the computational perspective, where patients are grouped according to their clinical patterns. Various unsupervised deep learning models are applied to learn representations for better subtyping. Morid et al. [20] proposed DeepPatient, a framework that consists of a three-layer stack of denoising autoencoders, to capture the aggregated historical medical codes. To capture the temporal information between medical events, Ruan et al. [21] applied recurrent neural network-based denoising autoencoder (RNN-DAE) to learn patient representation. Huang et al. [22] developed an outcome-aware deep clustering network, called DICE, that can encode the medical data and predict the outcome at the same time. DICE is able to learn cluster membership based on the historical data and serve as a pseudo label for unseen patients. Risk levels can therefore be assigned to new patients. Time-Aware sequential autoencoder encoder is proposed by Baytas et al. [23] to handle the irregular time intervals in the longitudinal patient records. Compared to standard RNN units, their time-aware RNN units can better capture the irregular time gaps between medical visits. Landi et al. [24] leverage Convolutional AutoEncoder (ConvAE) to generate patient embeddings for patient sub-typing. Their framework is scalable and helps to understand the etiology variation in heterogeneous subpopulations. Despite the promising result provided by these methods, they failed to incorporate the resource usage information, which is vital for risk stratification and analysis.

## 3 Model Architecture

In this section, we describe our proposed framework for learning efficient patient representation.

### 3.1 Notation

Patients in the claims data can be represented as a sequence of medical visits $v_1, v_2, v_3, \dots, v_i$ ordered by service date $t$. The $i^{th}$ medical visit $v_i$ contains a set of medical codes $\{c_1, c_2, \dots, c_j\} \subseteq |C|$, where $|C|$ refers to the vocabulary size of the medical codes. Claims data contains three types of medical visits, i.e., inpatient visit, outpatient visit and pharmacy visit. The visit type can be represented by the utilization code, where $IP$ indicates the inpatient visit, $OP$ indicates the outpatient visit, and $RX$ indicates the pharmacy visit.

### 3.2 Encoder and Decoder

The goal of our model is to learn an effective vector representation that encodes the patient's health condition and resource utilization pattern. This goal can be achieved by capturing patients' historical multimodal claim information, including medical codes, service dates, utilization types, and incurred medical expenditure. As shown in Figure 2, our proposed architecture contains two main parts: an encoder to model the claims data and a decoder to reconstruct the sequential information based on learned patient representation. Details about the model structure are shown below.

**Transformer-based Encoder.** Our encoder adapted the classic transformer architecture [10] as the building block, which contains a *Multi-Head Self-Attention layer*, an *Add & Normalize layer* and a *Multi-Level Feed-Forward layer*. Due to the multimodal input structure of the claims data, we first carefully designed the embedding layers for better compressing various information into medical visit vectors.

Inspired by the position embedding technique [25,26] and recent practices on variable encoding methods, we utilize an embedding map to convert the medical expenditure variable and service date variable into continuous embedding space. Specifically, given the observed expenditures in the dataset, we sort the values and discretize them into 100 sub-ranges with an equal number of observed expenditures in each sub-range. An expenditure variable can then be map to a vector $E_{cost} \in R^d$ through cost embedding layer, where $d$ is the dimension for variable embedding. For the service date variable, we subtract the service date from the medical enrollment start date and then the resulting integer is converted into a one-hot vector. Next, we embed the vector into latent space, $E_{date} \in R^d$, as follows,

$$E_{date/cost}(pos, 2i) = \sin\left(pos/10000^{2i/d}\right)$$
$$E_{date/cost}(pos, 2i+1) = \cos\left(pos/10000^{2i/d}\right).$$

For patients' demographics information, we first encode ages into several age groups (i.e., 2-5, 5-8, 8-12, etc.). The age group and gender are then sent to an embedding layer and represented by embedding matrix $E_{demo} \in R^{2*d}$. Similarly, the utilization embedding is obtained by first encoding the utilization token into a one-hot vector then passing through an embedding layer, $E_{util} \in R^d$.

Multiple diagnoses, procedures, and medication codes might exist in medical events and pharmacy events. For these variables, we first convert them into multi-hot vectors then pass them through the corresponding embedding layer, i.e., $E_{diag} \in R^{j*d}; E_{proc} \in R^{k*d}, E_{drug} \in R^{l*d}$.

Finally, given the above embedding matrices, we concatenate them in the new semantic space and obtain embedding matrix $E_t$, which contains multi-modal information at each time visit time t. Since the numbers of medical codes at different visit times are not identical, the dimensions of $E_t$ are varying. To address the dimension irregular issue, we adopt a max-pooling layer extract distinctive feature of embedding matrix $E_t$. The resulting vector is referred to $e_t$, which is then sent to the encoder. The transformer-based encoder can be described as follows,

$$pe, \widetilde{e_1}, \widetilde{e_2}, \dots, \widetilde{e_t} = TransLayer([E_{demo}; e_1, e_2, \dots, e_t])$$
$$e_t = maxpool(E_t)$$
$$E_t = [E_{diag}; E_{proc}; E_{drug}; E_{util}; E_{date}; E_{cost}],$$

where $TransLayer$ refers to the transformer encoder layer in [10], $pe, \widetilde{e_1}, \widetilde{e_2}, \dots, \widetilde{e_t}$ are the outputs of the encoder.

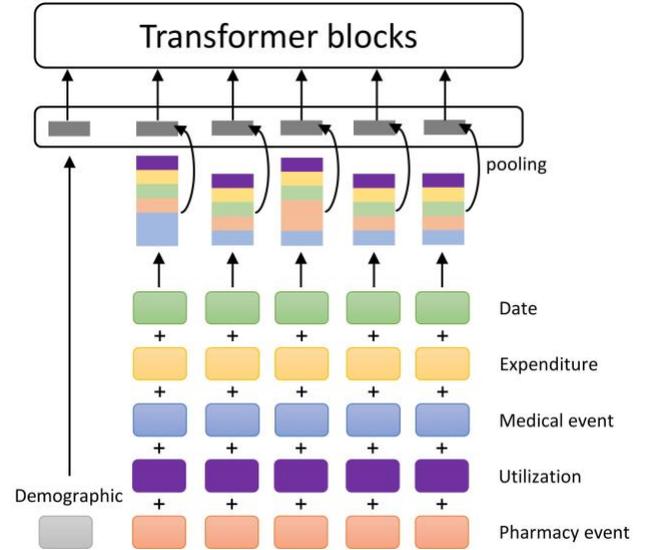

**Figure 3. Embedding layer for the transformer encoder of our TMAE architecture. All variables are converted into embeddings and aggregated as inputs via max-pooling for the transformer blocks.**

**Patient Embedding Layer.** The patient embedding, denoted as pe $\in R^d$, is obtained by the first output element of the encoder. The output of the encoder and the input of the decoder is fused at the patient embedding layer. The resulting patient embedding is fed as the input for the decoder. Through the two reconstruction tasks, patient embedding is forced to encode the clinical information as well as the resource usage

**Transformer-based Decoder**. We also adopt the transformer architecture for the decoder. The decoder receives the patient embedding, date embeddings, and utilization embeddings as inputs and reconstructs the expenditure, medical and pharmacy events in a non-autoregressive manner. The date and utilization embedding serve as queries to search for valuable information from the patient embedding to reconstruct the corresponding medical events and incurred cost. Considering the observation that medical events with high medical costs often indicate a severe medical condition and require more attention, we reconstruct the actual medical expenditure (i.e., in dollar amount) instead of the binary expenditure vector.

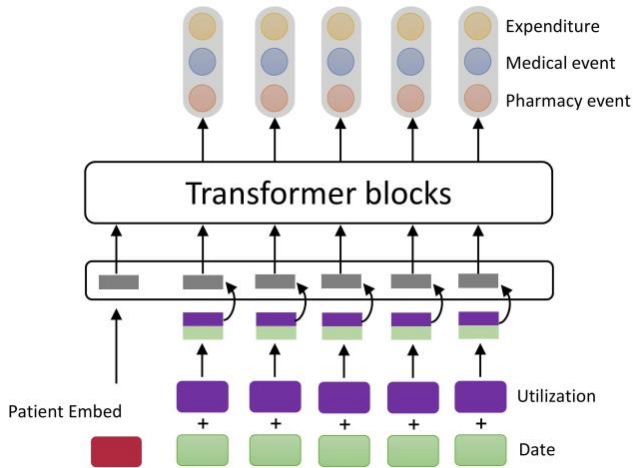

**Figure 4. Embedding layer for transformer decoder. Patient embedding, date embedding, and utilization embedding are used as inputs to reconstruct the expenditure, medical event, and pharmacy event**

### 3.3 Mitigate Sparsity Issue of Rare Medical Code

To alleviate the sparsity issue of the rare medical codes, we leverage pre-training techniques and Clinical Classifications Software (CCS) clinical groupers [27]. As shown in Figure 5, the medical code embeddings are built based on the medical code embedding and its corresponding category embedding. By concatenating these two embeddings, we obtain a vector containing both the medical code and its category information. Thus, even if some rare medical codes, such as Malignant Neoplasm of Brain (ICD-9 191.9), are extremely rare in the dataset, the corresponding medical category, such as Brain Cancer (CCS 35), can still provide valid information. This setting helps to mitigate the sparsity issue of rare medical codes. What is more, we pretrain both the code embeddings and category embeddings via med2vec [28]. The pretrained embedding matrix provides a better initialization value and improves the performance by inducing knowledge from the co-occurrence information.

### 3.4 Multi-objective Joint Training Scheme

For model training, we apply a joint training scheme consisting of two tasks, i.e., medical event reconstruction task and medical expenditure prediction task. The details are shown below.

**Task #1: Medical Event Reconstruction**. To encode the clinical information and the temporal patterns, we train the proposed architecture to reconstruct the sequential medical events (i.e., diagnosis, procedures, and medications). We use sigmoid cross-entropy as the loss function and denote it as $L_{code}$.

**Task #2: Medical Expenditure Prediction**. The medical expenditure prediction task aims at estimating the medical cost of each medical visit. One advantage of this task is that it helps the patient embedding to encode the resource usage information. Another advantage is that it forces the model to pay more attention to the high-cost visit as a misprediction of the high-cost visit will lead to a more significant penalty. Mean absolute error is chosen as the loss function and denoted as $L_{cost}$.

We combine two losses using loss coefficient lambda as follows, $L_{loss} = L_{cost} + \lambda\ L_{code}$. In practice, we tend to select a lambda value that can balance the two losses.

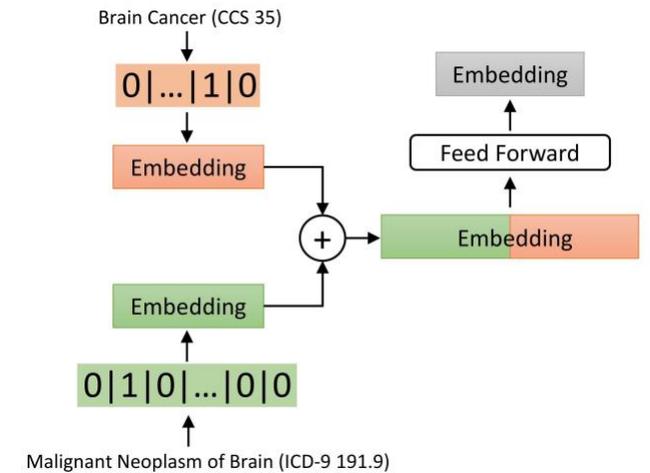

**Figure 5. Illustration of our embedding technique to mitigate the sparsity issue of rare medical code. The medical category embedding helps to provide valid information for rare medical codes.**

### 4 Experimental Setting

This section describes the details of our experimental setting. Several evaluation tasks are conducted on a large real-world pediatric dataset, and five baselines are compared with our proposed TMAE framework. Section 4.1 describes the data we used, the preprocessing steps, and the implementation details. In Section 4.2, we demonstrate the five baselines we used for model performance comparison.

## 4.1 Data Source and Preprocessing

The data source for this study is claims data from Partner For Kids (PFK). Partner for Kids (PFK) is one of the largest pediatric ACOs for Medicaid enrollees in central and southeastern Ohio. The dataset was obtained from a density sampled study that contains more than 600,000 enrollees' claims data. We extracted the de-identified encounter records from patients with two years of continuous eligibility from 2013 to 2014. In accordance with the Common Rule (45 CFR 46.102[f]) and the policies of Nationwide Children's Institutional Review Board, this study used a limited dataset and was not considered human subjects research and thus not subject to institutional review board approval.

As inputs, we use diagnosis code (International Classification of Diseases, Ninth Revision, Clinical Modification, ICD-9-CM), procedure code (ICD-9-CM procedure code, Current Procedural Terminology [CPT] and Healthcare Common Procedure Coding System [HCPCS]), medication codes (National Drug Codes [NDC]), type of claim (Claim Type Code), service date, and incurred expenditure (the actual paid amount to the care organization). During the study period, 6,416 unique diagnosis codes, 4,265 unique procedure codes, and 2,643 unique NDC codes were identified. CCS and NDC-Directory are adopted to group unique medical codes into approximately 1,500 categories.

The claim service dates were subtracted by the year's start date and converted to an integer range from 0 to 365. All negative paid amounts were converted to zero, and claims with empty service dates or missing patient ids were removed (less than 0.1% of such claims).

Models are implemented using the TensorFlow library [29]. The model dimensions are selected carefully to make sure all networks share a similar number of neurons. The number of transformer layers for both the encoder and decoder is set to one. The loss coefficient lambda is set to 2e-6 to balance the two losses. Our codes are publicly available on the following GitHub Page: https://github.com/1230pitchanqw/TMAE-.

## 4.2 Methods for Comparison

We applied five unsupervised representation learning models to encode claims data, including PCA, DeepPatient [20], RNN-DAE [21], TLSTM-AE [23], Conv-AE [24]. The following section describes the details for these models.

- **PCA**. PCA is one of the most widely used methods for dimension reduction, and it serves as the base representation learning model in the healthcare domain. We first aggregate the sequential claims data and fit a PCA model as our first baseline.
- **DeepPatient**. Proposed by Miotto et al. [20], DeepPatient consists of a three-layer feed-forward network as the encoder and a three-layer feed-forward network as the decoder. DeepPatient shows promising performance in auto diagnosis tasks and is used as one of our baselines.
- **RNN-DAE**. Recurrent Neural Networks are popular for modeling EHR as their ability to process sequence data. RNN-based models have been successfully applied to learn effective patient embedding for mortality prediction and heart failure detection [30]. We adopt the setting from [21] and use it as another baseline.
- **TLSTM-AE**. Instead of using a regular RNN unit, which is designed to handle data with constant elapsed times, TLSTM-AE leverages a time-aware unit that is able to handle the irregular time intervals in longitudinal patient records.
- **Conv-AE**. Convolutional Autoencoder networks are another set of neural networks designed to transform data into low dimensional space. In [24], Landi et al. use Conv-AE to encode patient trajectories into embeddings for stratification. We include the proposed Conv-AE as the last baseline.
- **TMAE**. Transformer-based Multi-modal AutoEncoder (TMAE) is our proposed model to learn patient representation from claims data. To evaluate the effectiveness of the components of our proposed architecture, including multi-objective training scheme and disease & category embedding concatenation, we implement two reduced models as shown below.
- **P-TMAE**. P-TMAE initializes medical code vectors directly without concatenating the corresponding category vectors.
- **C-TMAE**. C-TMAE removes the L-cost (i.e., Medical Expenditure Prediction Task) in the objective function.

## 5 Results

Our proposed model is compared with the five baselines on various downstream tasks. In this section, the tasks and the corresponding evaluation metrics are described in each subsection, followed by the result presentation and performance discussion.

## 5.1 Methods for Comparison

We first would like to evaluate the models' ability to encode clinical and resource usage information in the claims data. To this end, two clustering tasks using different patient cohorts that satisfy certain criteria are conducted. Specifically,

- For the first clustering task, analysis is performed using patients with the following four health conditions: asthma, diabetes, depression, and seizure (identified through the ICD-9 diagnosis code). Three hundred patients were randomly selected for each health condition, and a total of 1,200 patients are used to construct the cohort.
- For the second clustering task, the patient cohort consists of 1,100 asthma patients that have 10-12 asthma-related medical visits. Among these patients, 100 of them incurred high medical expenditure (i.e., spent more than $8000 annually), 500 of them incurred medium medical expenditure (i.e., spent more than $1000 but less than $2000 annually), and 500 of them incurred low medical expenditure (i.e., spent more than $100 but less than $500 annually).

The performance of algorithms in the two clustering tasks was evaluated using the Calinski-Harabasz Index [31] and Davies-Bouldin Score [32]. Calinski-Harabasz Index (C-H) and Davies-Bouldin Score (D-B) are two popular evaluation metrics for clustering tasks. The C-H can be used to evaluate the model performance when the ground truth labels are unknown. In this metric, a higher score indicates a model with better-defined clusters. The minimum D-B is zero, with a lower score indicating a better clustering result and, therefore, better model performance. We adopt Scikit-Learn's implementation of C-H and D-B, where the detailed definition and calculation can be found on [33].

Table 1. Stratification result evaluation via two clustering metrics: C-H index and D-B score.

|  |  | Task 1 | | Task 2 | |
| --- | --- | --- | --- | --- | --- |
|  |  | C-H | D-B | C-H | D-B |
| Baselines | PCA | 12.70 | 6.58 | 31.24 | 4.18 |
|  | Deep-Patient | 18.86 | 5.21 | 18.88 | 7.17 |
|  | RNN-DAE | 13.77 | 7.89 | 24.99 | 5.28 |
|  | TLSTM-AE | 12.10 | 8.02 | 18.42 | 6.09 |
|  | Conv-AE | 14.20 | 6.80 | 15.34 | 9.07 |
| Reduced Model | P-TMAE | 16.84 | 6.64 | 56.42 | 3.61 |
|  | C-TMAE | **36.54** | **4.20** | 39.65 | 4.07 |
| Proposed | TMAE | 34.15 | 4.41 | **64.17** | **3.51** |

In Table 1, the performance of PCA is the best among all baselines on Task 2. The superior performance is likely because the PCA successfully identifies correlated medical codes between three asthma cohorts and separates these cohorts by removing correlated features. On the other hand, neural network-based approaches, like DeepPatient, are less sensitive to redundant medical codes. These medical codes, such as outpatient visit procedure codes (CPT 99214), do not contribute to the clustering process, and yet they are still encoded in the latent space.

Compared to PCA and DeepPatient, the other three baselines are able to capture the temporal information between medical claims. However, we do not observe a clear performance boost. This observation indicates that modeling temporal information does not help to improve the model performance on these two clustering tasks.

We can observe that our proposed approach, including TMAE variants, significantly outperforms baselines with respect to Calinski-Harabasz Index and Davies-Bouldin Score. Specifically, C-TMAE largely improves the clustering results for patient cohort-1, and P-TMAE dramatically improves the model performance for patient cohort-2. These improvements demonstrate the effectiveness of our proposed sparsity alleviating technique and the multi-objective joint training scheme.

For clustering task 1, we visualize the patient embeddings through t-SNE in Figure 6. We can clearly observe four tight patient clusters that well-align with the corresponding health conditions from the figure. The visualization confirms that our TMAE is able to capture the semantic meaning of medical codes and separate patients based on their disease conditions.

In Figure 7, the clustering result is demonstrated through the visualization of the representation. Three visible separations are found in the figure, which indicates that TMAE can encode resource usage information and subtype asthma patients based on their utilization patterns. It is interesting to see that the high utilizers (color blue) are surrounded by medium utilizers (color red) and far away from the low utilizers (color yellow). This observation matches our intuition that high utilizers are more "similar" to medium utilizers than low utilizers.

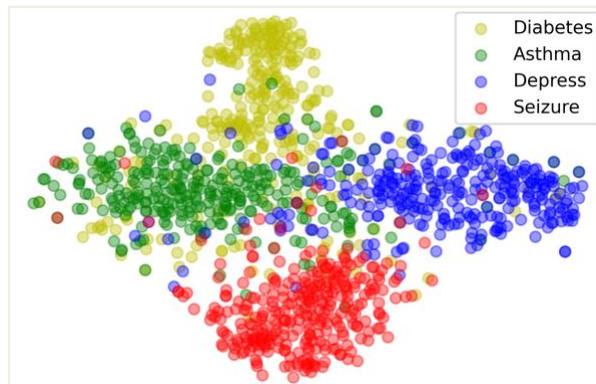

**Figure 6. TMAE learned patient embeddings visualization via t-SNE. Patients with different diseases are drawn in different colors.**

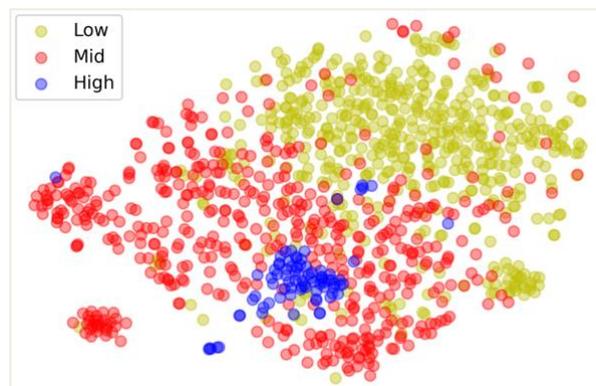

**Figure 7. TMAE learned patient embeddings visualization via t-SNE. Patients with resource usage patterns are drawn in different colors.**

## 5.1 Risk-Stratify Disease-Specific Population

In this subsection, we applied our method to stratify patients from two disease-specific cohorts, i.e., the autistic cohort and depression cohort. To determine the optimal number of clusters in each patient cohort, we adopt the Elbow method and the K-means clustering algorithm. The Elbow method investigates the total within-cluster sum of squares (WSS) as a function of the number of clusters and identifies the optimal number of clusters so that adding another cluster only marginally improves the total WSS.

As shown in Figure 8 (left), three clusters are identified in the autism cohort according to the elbow method. Figure 8 (right) displays the t-SNE plot of these three clusters, where patients belonging to different clusters are drawn in different colors. Three visually identifiable groups can be seen in the visualization. This observation indicates that our proposed TMAE can capture the hidden patterns of overlapping phenotypes and therefore can potentially be used for risk stratification.

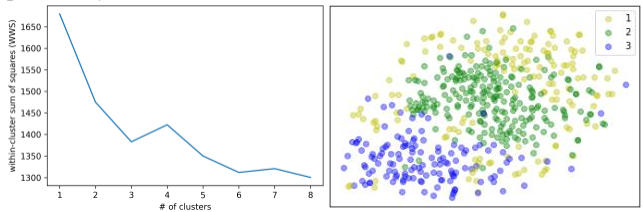

**Figure 8. Within-cluster Sum of Squares result (left) and visualization of subtyping (right) on autism population.**

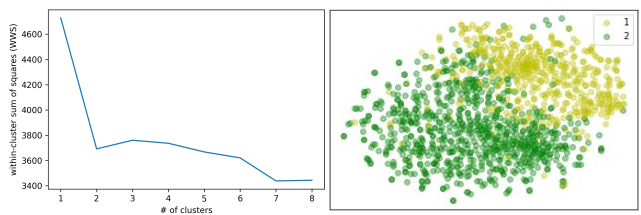

**Figure 9. Within-cluster Sum of Squares result (left) and visualization of subtyping (right) on depression population.**

Table 2 shows the demographic and resource burden of autism patients in the three identified subgroups. Each subgroup has more than 100 patients and depicts different characteristics. In particular, cluster -1 is characterized by low pharmacy cost; cluster-2 contains patients that are younger and cost less amount of money compare to the other two subgroups; cluster -3 shows signs of severe disease burden as the number of hospitalizations and pharmacy expenditures are higher than the other two subgroups.

Another stratification example is shown in Figure 9. Figure 9 (left) shows that two clusters are identified by the elbow method from the WSS plot. Figure 9 (right) present the t-SNE plot of the two clusters. We can observe that these two clusters are well-separated, indicating that different subtypes of depression patients can be identified using the TMAE learned patient representation.

Table 3 depict the statistic of patients in different clusters. Patients in cluster-1 show severe depression problems, identified by a significantly higher cost on medication (RX cost) and hospitalization rate (IP visit). On the other hand, patients in cluster-2 have less disease and resource burden. It is interesting to observe that the outpatient visit number in the mild depression group (i.e., cluster-2) is higher than the severe depression group (i.e., cluster-1). This observation might because patients who visit the hospital regularly are more likely to keep the depression condition on hold. It is also interesting to see that the female percentage in cluster-1 and cluster-2 diverse significantly.

Note that the main purpose of this study is to illustrate the potential of our proposed TMAE framework for risk stratification and risk analysis. Therefore, we do not discuss the clinical meaning behind the characteristic differences. We plan to address these findings by involving medical experts to review the clustering results in the future. In addition, due to the page limits, we only present the risk-stratification result on autism and depression population, more disease-specific stratification results, such as diabetes population stratification, can be found on our GitHub page.

Table 2: The demographic and resource burden of autism patients in different subgroups.

|  | Cluster-1 n=150 | Cluster-2 n=133 | Cluster-3 n=226 |
|---|---|---|---|
| Aver. Age | 10 | 6 | 10 |
| Female % | 19 | 20 | 18 |
| Aver. # OP visit | 9.3 | 12.2 | 8.1 |
| Aver. # IP visit | 0.02 | 0.008 | 0.05 |
| Median RX cost (Y1 / Y2) | $86/ $115 | $38/ $32 | $1139/ $1168 |
| Median Total cost (Y1 / Y2) | $1355/ $884 | $1221/ $1070 | $2087/ $1972 |

Table 3: The demographic and resource burden of depression patients in different subgroups.

|  | Cluster-1 n=481 | Cluster-2 n=796 |
|---|---|---|
| Aver. Age | 11 | 13 |
| Female % | 29 | 64 |
| Aver. # OP visit | 8.2 | 10.2 |
| Aver. # IP visit | 0.21 | 0.06 |
| Median RX cost (Y1 / Y2) | $ 1861/ $ 1662 | $ 164/ $ 149 |
| Median Total cost (Y1 / Y2) | $ 2650/ $ 2574 | $ 1508/ $ 1124 |

## 6  Limitation and Future Work

This study is limited in several ways. First, because ground truth labels of the stratification are unknown, we lack a quantitative measure of the fidelity of our subtyping with respect to the actual risk level. Although the clinical differences between subgroups are statistically significant, we lose some purity during the clustering process. We do not know how important this loss is, and in practice, some incorrect cases might be more important than the correct ones as they can hinder health intervention. Second, though our method is able to split the population into subgroups, the confidence for this measure is unknown. Different random seeds could result in a slightly different splitting. Our method did not provide confidence intervals for the patient split. Finally, interpretability is vital in the healthcare domain, and the black-box nature of our approach poses challenges for clinicians to understand the logic behind the subtyping.

Each of the above-mentioned limitations points to a good direction for future work. In the next step, we plan to quantify the fidelity of the subtyping by involving clinicians to review these subgroups manually. In addition, we are considering leveraging the bootstrapping method [34] for confidence interval estimation. Combining the previous two steps, we would like to cooperate with the domain experts to set up risk-stratification guidelines for using our framework on certain disease-specific populations. The guidelines can mitigate the faithful and ethical concerns raised by the black-box nature of our framework.

## 7  Conclusion

In this study, we proposed TMAE, a Transformer-based Multimodal AutoEncoder that can learn efficient patient representation from claims data for various risk-related applications. We conducted experiments on a large real-world dataset to evaluate the performance of our framework. These intensive experiments show promising results and demonstrate that our method is superior compared to various baselines. Specifically, our model achieved higher clustering score values on the two clustering tasks, which demonstrate the effectiveness of our framework for capturing the clinical and utilization patterns. In addition, we have successfully risk-stratified subgroups of patients for the disease-specific population. These subgroups show distinguishing resource usage characteristics, which enable the care organization to tailor management plans for better care intervention. In summary, our proposed framework can be successfully applied for risk stratification and risk analysis, which paves the way for improved care intervention and care management.

## ACKNOWLEDGMENTS

The authors would like to thank Brad Stamm and Wang Ling from Partners for Kids for preparing the claims data and valuable discussion on data preprocessing.